\documentclass[10pt,twocolumn,letterpaper]{article}

\usepackage{cvpr}              

\usepackage[dvipsnames]{xcolor}

\usepackage{pifont}
\usepackage{soul}
\usepackage{colortbl}
\usepackage{comment}
\usepackage{wrapfig, lipsum, booktabs}
\usepackage{makecell}
\usepackage{enumitem}
\usepackage{amsmath}
\usepackage{nicematrix}
\usepackage{adjustbox}

\definecolor{ForestGreen}{RGB}{0, 179, 45}
\newcommand{\tcb}{\textcolor{black}}
\newcommand{\tcr}{\textcolor{black}}
\newcommand{\tco}{\textcolor{black}}
\newcommand{\tcg}{\textcolor{black}}

\mathchardef\mhyphen="2D

\newcommand{\ieno}{\textit{i}.\textit{e}.}

\newcommand{\sota}{{state-of-the-art}}
\newcommand{\tv}{{\texttt{T2V}}}
\newcommand{\ti}{{\texttt{T2I}}}
\newcommand{\ours}{{PixelDance}}

\definecolor{cvprblue}{rgb}{0.21,0.49,0.74}
\usepackage[pagebackref,breaklinks,colorlinks,citecolor=cvprblue]{hyperref}

\def\paperID{11972} 
\def\confName{CVPR}
\def\confYear{2024}

\title{Make Pixels Dance: High-Dynamic Video Generation}

\author{Yan Zeng$^{*}$ ~~~Guoqiang Wei$^{*}$ ~~~Jiani Zheng \\ 
Jiaxin Zou  ~~~Yang Wei ~~~Yuchen Zhang ~~~Hang Li \\ \\
ByteDance Research \\
{\tt\small $^{*}$ Equal Contribution} \\
{\tt\small \{zengyan.yanne, weiguoqiang.9, lihang.lh\}@bytedance.com} \\
\href{https://makepixelsdance.github.io}{\texttt{https://makepixelsdance.github.io}}
}

\begin{document}

\twocolumn[{%
\renewcommand\twocolumn[1][]{#1}%
\maketitle
\begin{center}
    \centering
    \vspace{-0.6cm}
    \captionsetup{type=figure}
    \includegraphics[width=0.99\textwidth]{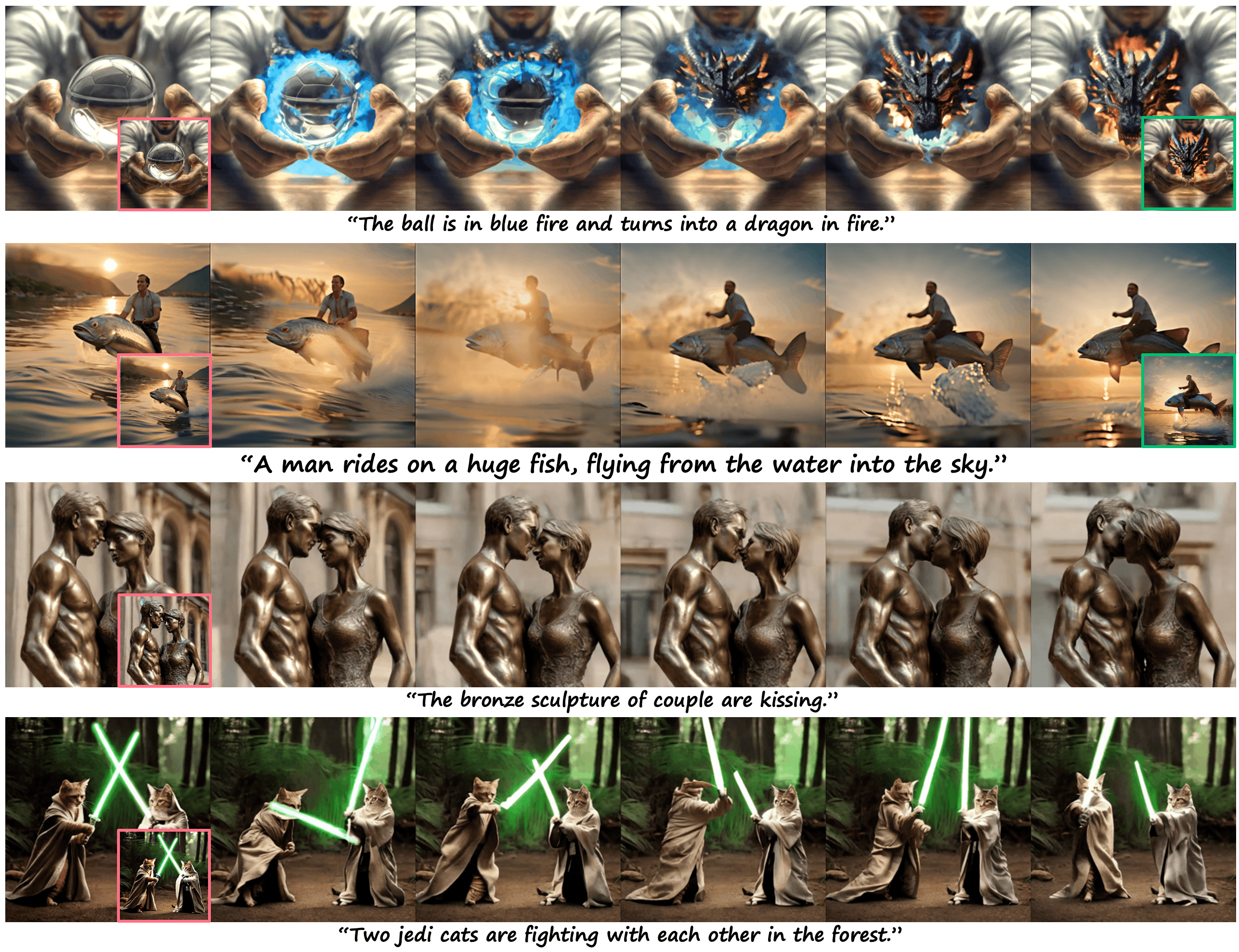}
    \captionof{figure}{Generation results of \textbf{\ours} given text, first frame instruction highlighted in red box (and last frame instruction in green). Six frames sampled from a 16-frame clip are displayed. Human faces presented in this paper are synthesized using text-to-image models.
    }
    \label{fig: teaser}
\end{center}%
}]

\begin{abstract}

Creating high-dynamic videos such as motion-rich actions and sophisticated visual effects poses a significant challenge in the field of artificial intelligence. Unfortunately, current state-of-the-art video generation methods, primarily focusing on text-to-video generation, tend to produce video clips with minimal motions despite maintaining high fidelity. We argue that relying solely on text instructions is insufficient and suboptimal for video generation. In this paper, we introduce \textbf{\ours}, a novel approach based on diffusion models that incorporates image instructions for both the first and last frames in conjunction with text instructions for video generation. Comprehensive experimental results demonstrate that \ours trained with public data exhibits significantly better proficiency in synthesizing videos with complex scenes and intricate motions, setting a new standard for video generation. 

\end{abstract}

\section{Introduction}

Generating high-dynamic videos with motion-rich actions, sophisticated visual effects, natural shot transitions, or complex camera movements, has long been a lofty yet challenging goal in the field of artificial intelligence. Unfortunately, most existing video generation approaches focusing on text-to-video generation~\cite{VDM,zhou2022magicvideo,ge2023preserve} are still limited to synthesize simple scenes, and often falling short in terms of visual details and dynamic motions. Recent state-of-the-art models have significantly enhanced text-to-video quality by incorporating an image input~\cite{GEN2,li2023videogen,wang2023modelscope}, which provides finer visual details for video generation. Despite the advancements, the generated videos frequently exhibit limited motions as shown in Figure~\ref{fig: gen2_cases}. This issue becomes particularly severe when the input images depict \tco{out-of-domain content unseen in training data}, indicating a key limitation of current technologies.

To generate high-dynamic videos, we propose a novel video generation approach that incorporates image instructions for both the first and last frames of a video clip, in addition to text instruction. The image instruction for the first frame depicts the major scene of the video clip. The image instruction for the last frame, which is {\it optionally} used in training and inference, delineates the ending of the clip and provides additional control for generation. The image instructions enable the model to construct intricate scenes and actions. Moreover, our approach can create \tcg{longer} videos, in which case the model is applied multiple times and the last frame of the preceding clip serves as the first frame instruction for the subsequent clip.

The image instructions are more direct and accessible compared to text instructions. We use ground-truth video frames as the instructions for training, which is easy to obtain. In contrast, recent work has proposed the use of highly descriptive text annotations~\cite{dalle3} to better follow text instructions. Providing detailed textual annotations to precisely describe both the frames and the motions of videos is not only costly to collect but also difficult to learn for the model. To understand and follow complex text instructions, the model needs to significantly scale up. The use of image instructions overcome these challenges together with text instructions. Given the three instructions in training, the model is able to focus on learning the dynamics of video content, and in inference the model can better generalize the learned dynamics knowledge to out-of-domain instructions.

Specifically, we present \textbf{\ours}, a latent diffusion model based approach to video generation, conditioned on \texttt{<text,first frame,last frame>} instructions. The text instruction is encoded by a pre-trained text encoder and is integrated into the diffusion model with cross-attention. The image instructions are encoded with a pre-trained VAE encoder \cite{rombach2022high} and concatenated with either perturbed video latents or Gaussian noise as the input to the diffusion model. 
In training, we use the (ground-truth) first frame to enforce the model to strictly adhere to the instruction, maintaining continuity between consecutive video clips. In inference, this instruction can be conveniently obtained from \ti~models \cite{rombach2022high} or directly provided by users.

Our approach is unique in its way of using the last frame instruction. We intentionally avoid encouraging the model to replicate the last frame instruction exactly since it is challenging to provide a perfect last frame in inference, and the model should accommodate user-provided coarse drafts for guidance. Such kind of instruction can be readily created by the user using basic image editing tools.

To this end, we develop three techniques. First, in training, the last frame instruction is randomly selected from the last three (ground-truth) frames of a video clip. Second, we introduce noise to the instruction to mitigate the reliance on the instruction and promote the robustness of model. Third, we randomly drop the last frame instruction with a certain probability, e.g. 25\%, in training. Correspondingly, we propose a simple yet effective sampling strategy for inference. During the first $\tau$ denoising steps, the last frame instruction is utilized to guide video generation towards the desired ending status. Then, during the remaining steps, the instruction is dropped, allowing the model to generate more temporally coherent video. The impact of last frame instruction can be adjusted by $\tau$.

Our model's ability of leveraging image instructions enables more effective use of public video-text datasets, such as WebVid-10M~\cite{bain2021WEBVID} \tcb{which only contains coarse-grained descriptions with loose correlation to videos~\cite{singer2022make}, and lacks of content in diverse styles (\textit{e.g.}, comics and cartoons).} 
Our model with only 1.5B parameters, trained mainly on WebVid-10M, achieves \sota~performance on multiple scenarios. First, given text instruction only, \ours~leverages \ti~models to obtain the first frame instruction to generate videos, reaching FVD scores of 381 and 242.8 on MSR-VTT \cite{xu2016msr-vtt} and UCF-101 \cite{soomro2012UCF} respectively. With the text and first frame instructions (the first frame instruction can also be provided by users), \ours~is able to generate more motion-rich videos compared to existing models. Second, \ours~can generate continuous video clips, outperforming existing long video generation approaches~\cite{he2022LVDM,ge2022TATS} in temporal consistency and video quality. Third, the last frame instructions are shown to be a critical component for creating intricate out-of-domain videos with complex scenes and/or actions, as shown in Figure~\ref{fig: teaser}. 
\tcr{Overall, by actively interacting with \ours, we create the first three-minute video with a clear storyline at various complex scenes and characters hold consistent across scenes. }

Our contributions can be summarized as follows:
\begin{itemize}
\item We propose a novel video generation approach based on diffusion model, \textbf{\ours}, which incorporates image instructions for both the first and last frames in conjunction with text instruction.

\item We develop training and inference techniques for \ours, \tcg{which not only} effectively enhances the quality of generated videos, but also provides users with more control over the video generation process.

\item Our model trained on \tcg{public} data demonstrates remarkable performance in high-dynamic video generation with complex scenes and actions, setting a new standard for video generation.
\end{itemize}

\section{Related Work}

\subsection{Video Generation}

\begin{figure}[t]
  \centering
  \includegraphics[width=0.475\textwidth]{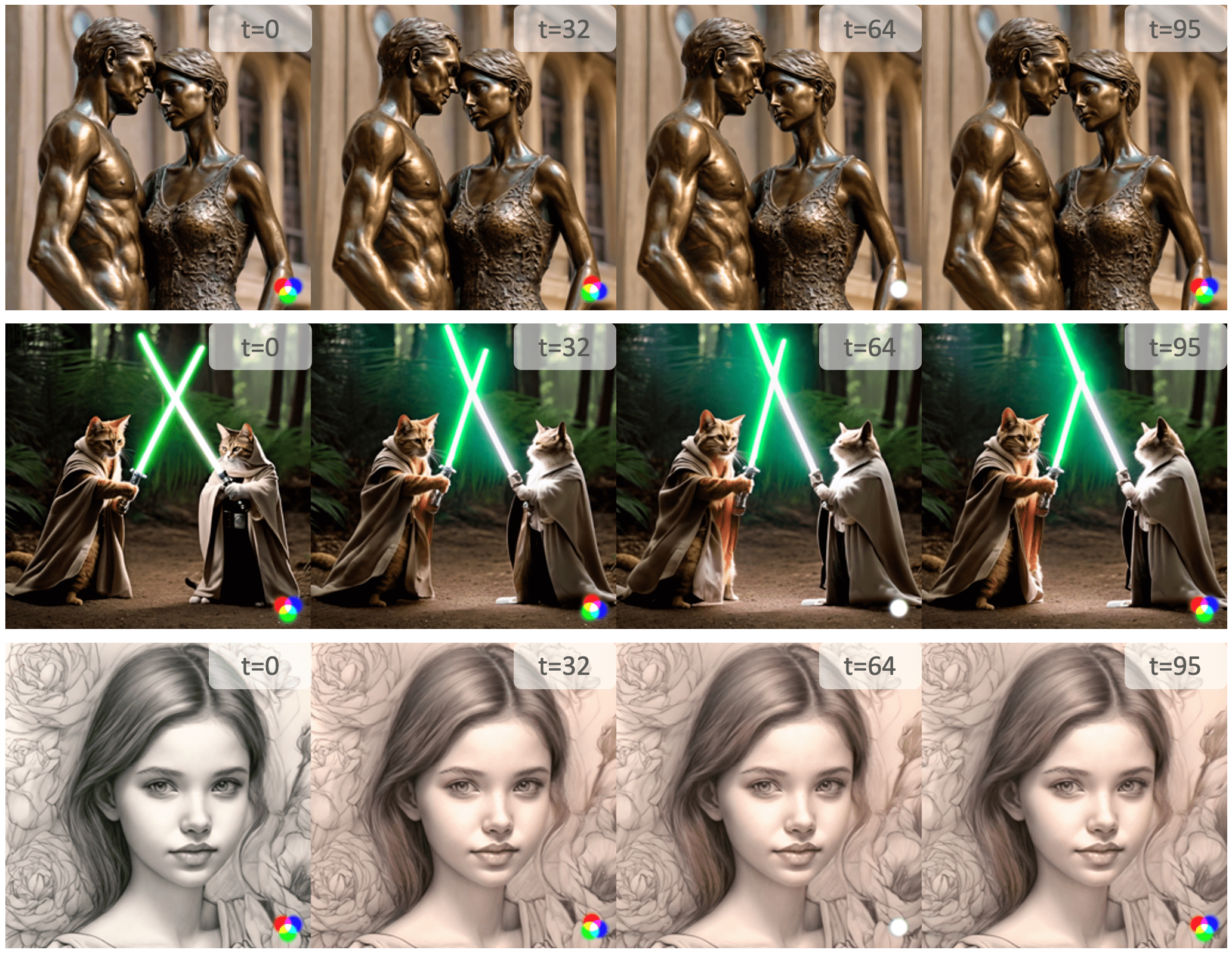}
  \caption{Videos generated by \sota~video generation model~\cite{GEN2}, compared with our results given the same text prompts and image conditions in Figure \ref{fig: teaser} and Figure \ref{fig: vis_rotate}. 
  }
  \label{fig: gen2_cases}
\end{figure}

Video generation has long been an attractive and essential research topic~\cite{ranzato2014video, vondrick2016generating,ge2022TATS}. Previous studies have resorted to different types of generative models such as GANs~\cite{goodfellow2014generative,li2018video,pan2018create,tian2021good} and Transfomers with VQVAE~\cite{yan2021videogpt,ge2022long,hong2022cogvideo}. Recently, diffusion models have significantly advanced the progress of photorealistic text-to-image generation~\cite{saharia2022photorealistic,balaji2022ediffi}, which exhibit robustness superior to GANs and require fewer parameters compared to transformer-based counterparts. Latent diffusion models~\cite{rombach2022high} are proposed to reduce the computational burden by training a diffusion model in a compressed lower-dimensional latent space. For video generation, previous studies typically add temporal convolutional layers and temporal attention layers to the 2D UNet of a pre-trained text-to-image diffusion models~\cite{singer2022make,he2022LVDM,zhou2022magicvideo,gu2023seer,wang2023modelscope,wang2023videocomposer,luo2023videofusion,ge2023preserve}. Although these advancements have paved the way for the generation of high-resolution videos through the integration of super-resolution modules~\cite{lu2023EGSVR}, the videos produced are characterized by simple, minimal motions as shown in Figure~\ref{fig: gen2_cases}. 

Recently, the field of video editing has witnessed remarkable progress \cite{yin2023dragnuwa,zhang2023magicavatar,molad2023dreamix}, particularly in terms of content modification while preserving the original structure and motion of the video, for example, modifying a cattle to a cow~\cite{gen1, tuneavideo}. Despite these achievements, the necessity to search for an appropriate reference video for editing is time-consuming. Furthermore, this approach inherently constrains the scope of creation, as it precludes the possibility of synthesizing entirely novel content (\textit{e.g.}, a polar bear walking on the 
Great Wall) that may not exist in any reference video.

\subsection{Long Video Generation}

Long video generation is a more challenging task which requires seamless transitions between successive video clips and long-term consistency of the scene and characters. There are typically two approaches: 1) autoregressive methods~\cite{VDM, villegas2023phenaki,harvey2022flexible} employ a sliding window to generate a new clip conditioned on the previous clip. 2) hierarchical methods~\cite{ge2022long, he2022latent, harvey2022flexible,yin2023nuwa} generate sparse frames first, then interpolate intermediate frames. However, the autoregressive approach is susceptible to quality degradation due to error cumulation over time. As for the hierarchical method, it needs long videos for training, which are difficult to obtain due to frequent shot changes in online videos. Besides, generating temporally coherent frames across larger time interval exacerbates the challenges, which often leads to low-quality initial frames, making it hard to achieve good results in later stages of interpolation. In this paper, \ours~generates continuous video clips in the auto-regressive way and exhibits superior performance in synthesizing long-term consistent frames compared to existing models. Concurrently, we advocate for active user engagement with the generation process, akin to a film director's role, to ensure that the produced content closely aligns with the user's expectation.

\section{Method}
\label{sec: method}

Existing models in text-to-video~\cite{VDM,zhou2022magicvideo,ge2023preserve} and image-to-video generation~\cite{GEN2,li2023videogen,wang2023modelscope} often produce videos characterized by simple and limited movements. In this paper, we attempt to enable the model to focus on learning the dynamics of video contents, to generate videos with rich motions. We present a novel approach that integrates image instructions for both the first and last frames in conjunction with text instruction for video generation\tcr{, and we effectively utilize public video data for training}. We will elaborate on the model architecture in Sec. \ref{sec: condition vdm}, and then introduce the training and inference techniques tailored for our approach in Sec. \ref{sec: infer}.

\subsection{Model Architecture}
\label{sec: condition vdm}

\begin{figure}[t]
  \centering
  \includegraphics[width=0.475\textwidth]{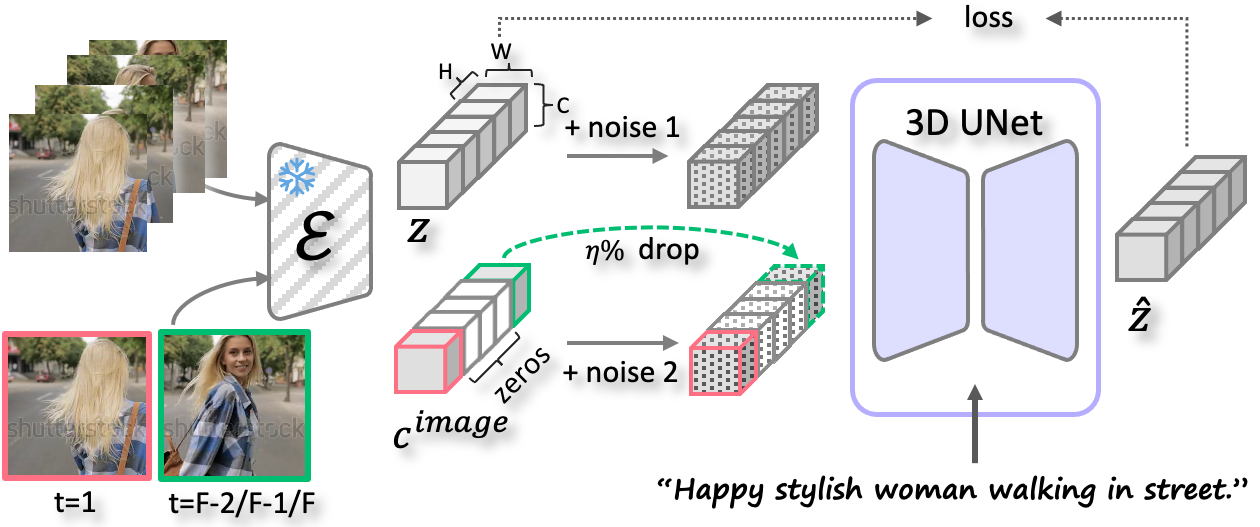}
  \caption{
  Illustration of the training procedure of \ours. The original video clip and image instructions (in red and green boxes) are encoded into $\mathbf{z}$ and $\mathbf{c}^{image}$, which are then concatenated along the channel dimension after perturbed with different noises. 
  }
  \label{fig: model}
\end{figure}

\paragraph{Latent Diffusion Architecture} We adopt latent diffusion model~\cite{rombach2022high} for video generation. Latent diffusion model is trained to denoise from a perturbed input in the latent space of a pre-trained VAE, in order to reduce the computational burden. We take the widely used 2D UNet \cite{ronneberger2015u} as diffusion model, which is constructed with \tcb{a series of spatial downsampling layers followed by a series of spatial upsampling layers} with inserted skip connections. Specifically, it is built with two basic blocks, \ieno, 2D convolution block and 2D attention block. We extend the 2D UNet to 3D \tcb{variant} with inserting temporal layers~\cite{VDM}, where 1D convolution layer along temporal dimension after 2D convolution layer, and 1D attention layer along temporal dimension following 2D attention layer. The model can be trained jointly with images and videos to maintain high-fidelity generation ability \tcb{on spatial dimension}. The 1D temporal operations are disabled for image input. We use bi-directional self-attention in all \tcb{temporal} attention layers. We encode the text instruction using a pre-trained CLIP text encoder~\cite{radford2021CLIP}, and the embedding $\mathbf{c}^{text}$ is injected through cross-attention layers in the UNet with hidden states as queries and $\mathbf{c}^{text}$ as keys and values.

\paragraph{Image Instruction Injection}
We incorporate image instructions for both the first and last frames in conjunction with text instruction. We utilize ground-truth video frames as the instructions in training, which is easy to obtain. Given the image instructions on the first and last frame, denoted as $\{ \mathbf{I}^{first}, \mathbf{I}^{last} \}$, we first encode them into the 
input space of diffusion models using VAE, result in $\{ \mathbf{f}^{first}, \mathbf{f}^{last} \}$ where $\mathbf{f}\in \mathbb{R}^{C\times H\times W}$. To inject the instructions without loss of the temporal position information, the final image condition is then constructed as:
\begin{align}
\mathbf{c}^{image} = [\mathbf{f}^{first}, \texttt{PADs}, \mathbf{f}^{last}]\in \mathbb{R}^{F \times C\times H \times W},
\label{eq: cat_condition}
\end{align}
where $\texttt{PADs} \in \mathbb{R}^{\left(F-2\right) \times C \times H \times W} $. \tcr{The condition $\mathbf{c}^{image}$ is then concatenated with noised latent $\mathbf{z}_t$ along the channel dimension, which is taken as the input of diffusion models.}

\subsection{Training and Inference}
\label{sec: infer}

\tcb{The training procedure is illustrated in Figure \ref{fig: model}.}
For the first frame instruction, we employ the ground-truth first frame \tcb{for training, making the model adhere to the first frame instruction strictly in inference}. In contrast, we intentionally avoid encouraging the model to replicate the last frame instruction exactly. 
During inference, the \tco{ground-truth} last frame is unavailable in advance, the model needs to accommodate user-provided coarse drafts for guidance to generate temporally coherent videos. To this end, we introduce three techniques. First, we randomly select an image from the last three ground-truth frames of a clip to serve as the last frame instruction for training. 
\tcb{Second, to promote robustness, we perturb the encoded latents $\mathbf{c}^{image}$ of image instructions with noise}.

Third, during training, we randomly drop the last frame instruction with probability $\eta$, by replacing \tcb{the corresponding latent}
with zeros. Correspondingly, we propose a simple yet effective inference technique. During inferene, in the first $\tau$ out of the total $T$ denoising steps, the last frame instruction is applied to guide the video generation towards desired ending status, and it is dropped in the subsequent steps to generate more plausible and temporally consistent videos:
\begin{align}
\tilde{\mathbf{x}}_\theta = 
\begin{cases}
\hat{{\mathbf{x}}}_\theta( \mathbf{z}_t, \mathbf{f}^{first}, \mathbf{f}^{last}, \mathbf{c}^{text} ), & \text{if } t < \tau \\
\hat{{\mathbf{x}}}_\theta( \mathbf{z}_t, \mathbf{f}^{first}, \mathbf{c}^{text} ), & \text{if } \tau \leq t \leq T \\
\end{cases}
.
\label{eq: drop_last_frame}
\end{align}
$\tau$ determines the strength of model dependency on last frame instruction, adjusting $\tau$ will enable various applications. For example, our model can generate high-dynamic videos without last frame instruction (\ieno, $\tau=0$). 
Additionally, we apply the classifier-free guidance \cite{ho2022classifierfree} in inference, which mixes the score estimates of the model conditioned on text prompts and without text prompts.

\section{Experiments}

\subsection{Implementation Details}
\label{sec: details}

Following previous work, we train the video diffusion model on WebVid-10M \cite{bain2021WEBVID}, which contains about 10M short video clips with an average duration of 18 seconds, predominantly in the resolution of 336 × 596. Each video is associated with a paired text which generally offers a coarse description weakly correlated with the video content. Another nuisance issue of WebVid-10M lies in the watermarks placed on all videos, which leads to the watermark's presence in all generated videos. Thus, we expand our training data with other self-collected $500$K watermark-free video clips depicting real-world entities such as humans, animals, objects, and landscapes, paired with coarse-grained textual descriptions. Despite comprising only a modest proportion, we surprisingly find that combining this dataset with WebVid-10M for training ensures that \ours~is able to generate watermark-free videos if the image instructions are free of watermarks.

\ours~is trained jointly on video-text dataset and image-text dataset. For video data, we randomly sample 16 consecutive frames with 4 fps per video. Following previous work \cite{ho2022imagenvideo}, we adopt LAION-400M \cite{schuhmann2021laion} as image-text dataset. Image-text data are utilized every 8 training iterations. The weights of pre-trained text encoder and VAE model are frozen during training. We employ DDPM~\cite{ho2020ddpm} with $T=1000$ time steps for training. A noise corresponding to 100 time steps is introduced to the image instructions $\mathbf{c}^{image}$. We first train the model at resolution of 256$\times$256, with batch size of 192 on 32 A100 GPUs for 200K iterations, which is utilized for quantitative evaluations. This model is then finetuned for another 50K iterations with higher resolution.  \tcb{We incorporate $\epsilon$-prediction \cite{ho2020ddpm} as training objective.}

\begin{table}[ht]
\begin{center}
\small
\captionof{table}{
Zero-shot \tv~performance comparison on MSR-VTT. All methods generate video with spatial resolution of 256$\times$256. Best in bold.
}
\begingroup
\begin{adjustbox}{center}
\resizebox{0.49\textwidth}{!}{ 

\setlength{\tabcolsep}{3pt}
\begin{tabular}{@{}lccccc@{}}

\toprule
Method & 
\#data &
\#params. &
CLIPSIM($\uparrow$) &
FVD($\downarrow$) \\

\midrule

CogVideo (En) \cite{hong2022cogvideo} & 5.4M & 15.5B & 0.2631 & 1294 \\
MagicVideo \cite{zhou2022magicvideo} & 10M & - & - & 1290 \\
LVDM \cite{he2022LVDM} & 2M & 1.2B & 0.2381 & 742  \\
Video-LDM \cite{blattmann2023alignVideo-LDM} & 10M & 4.2B & 0.2929 & - \\ 
InternVid \cite{wang2023internvid} & 28M & - & 0.2951 & - \\
ModelScope \cite{wang2023modelscope} & 10M & 1.7B & 0.2939 & 550 \\
Make-A-Video \cite{singer2022make} & 20M & 9.7B & 0.3049 & - \\
Latent-Shift \cite{an2023latentshift} & 10M & 1.5B & 0.2773 & - \\
VideoFactory \cite{wang2023videofactory} & - & 2.0B & 0.3005 & - \\

\midrule
\ours & 10M & 1.5B & \textbf{0.3125} & \textbf{381} \\

\bottomrule

\end{tabular}

}
\end{adjustbox}
\endgroup
\label{table: msrvtt}
\end{center}
\end{table}

\begin{table}[ht]
\begin{center}
\small
\captionof{table}{
Zero-shot \tv~ performance comparison on UCF-101. All methods generate video with spatial resolution of 256$\times$256. Best in bold.
}
\begingroup
\begin{adjustbox}{center}
\resizebox{0.49\textwidth}{!}{ 
\setlength{\tabcolsep}{3pt}
\begin{tabular}{@{}lcccccc@{}}

\toprule
Method & 
\#data &
\#params. &
IS($\uparrow$) &
FID($\downarrow$) &
FVD($\downarrow$) \\

\midrule

CogVideo (En) \cite{hong2022cogvideo} & 5.4M & 15.5B & 25.27 & 179.00 & 701.59 \\
MagicVideo \cite{zhou2022magicvideo} & 10M & - & - & 145.00 & 699.00 \\
LVDM \cite{he2022LVDM} & 2M & 1.2B & - & - & 641.80 \\ 
InternVid \cite{wang2023internvid} & 28M & - & 21.04 & 60.25 & 616.51 \\
Video-LDM \cite{blattmann2023alignVideo-LDM} & 10M & 4.2B & 33.45 & - & 550.61 \\
ModelScope \cite{wang2023modelscope} & 10M & 1.7B & - & - & 410.00 \\
VideoFactory \cite{wang2023videofactory} & - & 2.0B & - & - & 410.00 \\
Make-A-Video \cite{singer2022make} & 20M & 9.7B & 33.00 & - & 367.23 \\
VidRD \cite{gu2023VidRD} & 5.3M & - & 39.37 & - & 363.19 \\
Dysen-VDM \cite{fei2023DysenVDM} & 10M & - & 35.57 & - & 325.42 \\

\midrule
\ours & 10M & 1.5B & \textbf{42.10} & \textbf{49.36} & \textbf{242.82} \\ 

\bottomrule

\end{tabular}
}
\end{adjustbox}
\endgroup
\label{table: ucf_short}
\end{center}
\end{table}

\begin{figure}[t]
  \centering
  \includegraphics[width=0.475\textwidth]{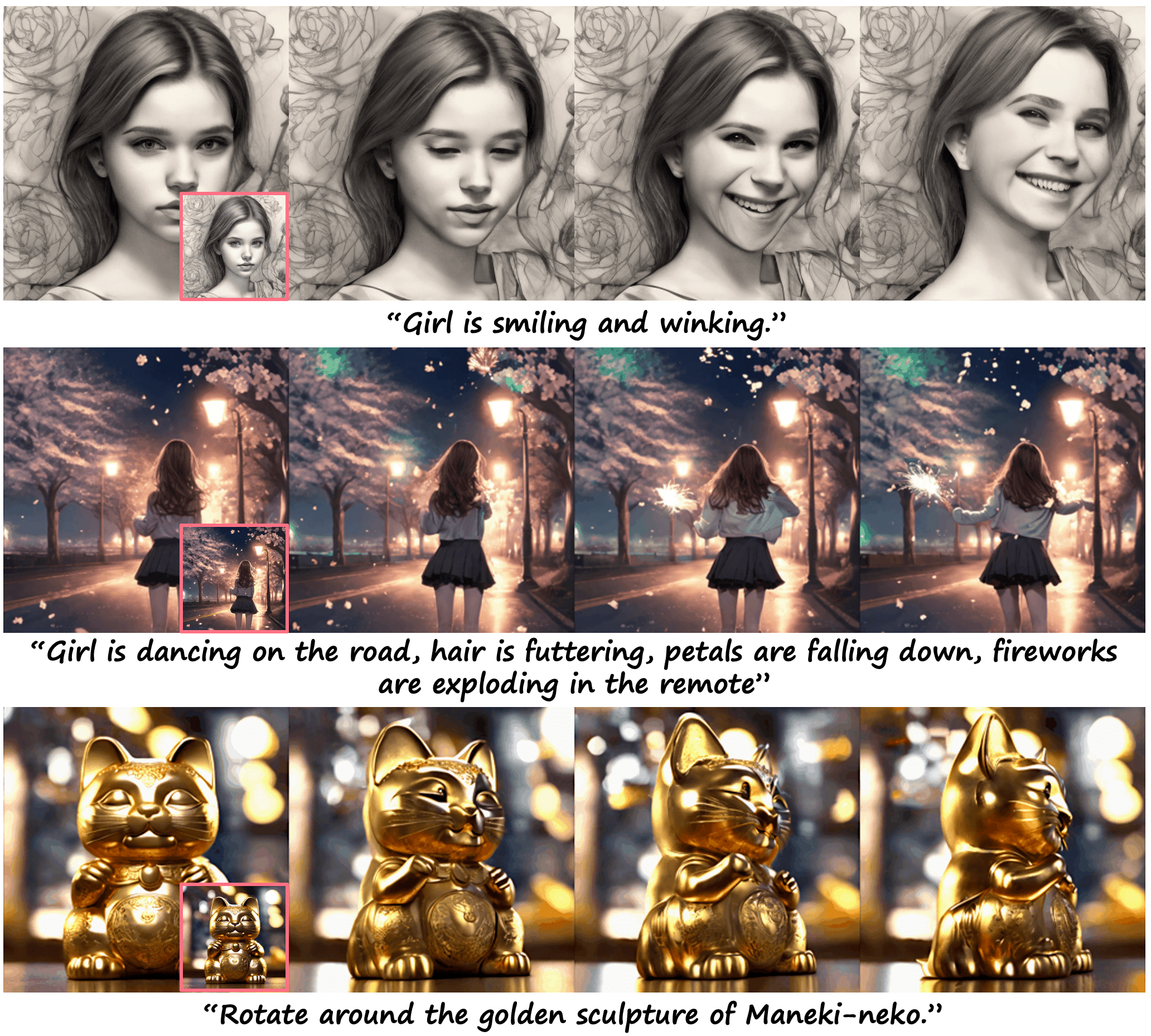}
  \caption{
  Illustration of video generation conditioned on the text and first frame instructions. Please refer to the Supplementary for more examples.
  }
  \label{fig: vis_rotate}
\end{figure}

\subsection{Video Generation}
\label{sec: exp_short}

\subsubsection{Quantitative Evaluation}
\label{sec: quant_eva}

We evaluate zero-shot video generation performance of our \ours~on MSR-VTT \cite{xu2016msr-vtt} and UCF-101 \cite{soomro2012UCF} datasets, following previous work \cite{hong2022cogvideo,zhou2022magicvideo,he2022LVDM,blattmann2023alignVideo-LDM}. MSR-VTT is a video retrieval dataset with descriptions for each video, while UCF-101 is an action recognition dataset with 101 action categories.
\textit{To make a comparison with previous \tv~approaches which are conditioned on text prompts only, we also evaluate only with text instructions.} Specifically, we utilize off-the-shelf \ti~Stable Diffusion V2.1 \cite{rombach2022high} to obtain the first frame instructions, and generate videos given the text and first frame instructions. Following previous work \cite{wang2023videofactory,fei2023DysenVDM}, \tcb{we randomly select one prompt per example to generate 2990 videos in total for evaluate,} and report the Fr\'echet Video Distance (FVD) \cite{unterthiner2018FVD} and CLIP-similarity (CLIPSIM) \cite{wu2021CLIPSIM} on MSR-VTT dataset. \tcb{For UCF-101 dataset, we construct descriptive text prompts per category and generate about 10K videos}, and compare with previous works in terms of the widely-used Inception Score (IS) \cite{saito2020IS},  Fr\'echet Inception Distance (FID) \cite{heusel2017FID} and FVD, following previous work \cite{wang2023videofactory,fei2023DysenVDM}. Both FID and FVD measure the distribution distance between generated videos and the ground-truth data. IS assesses the quality of generated videos and CLIPSIM estimates the similarity between the generated videos and corresponding texts.

Zero-short evaluation results on MSR-VTT and UCF-101 are presented in Table~\ref{table: msrvtt} and Table~\ref{table: ucf_short}, respectively. Compared to other \tv~approaches on the MSR-VTT, \ours~achieves \sota~result in terms of FVD and CLIPSIM, demonstrating its remarkable ability to generate high-quality videos with better alignment to text prompts. Notably, \ours~achieves an FVD score of 381, which substantially surpasses the previous \sota~ModelScope \cite{wang2023modelscope}, with an FVD of 550. On UCF-101 benchmark, \ours~outperforms other models across various metrics, including IS, FID and FVD.

\subsubsection{Qualitative Analysis} 
\label{sec: analysis}

\begin{figure}[t]
  \centering
  \includegraphics[width=0.475\textwidth]{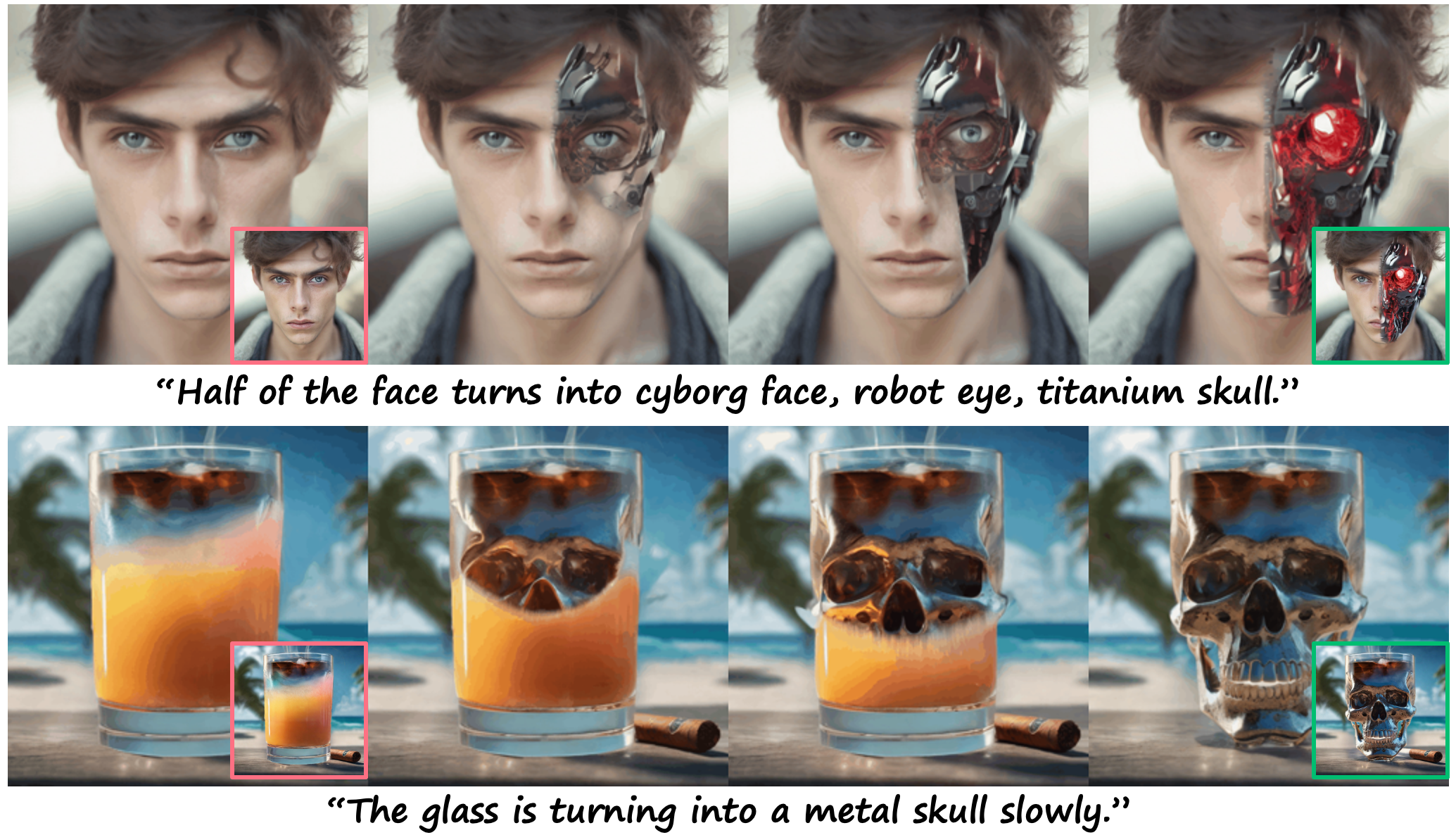}
  \caption{
  Illustration of complex video generation conditioned on the text, first frame and last frame instructions.  Please refer to the Supplementary for more examples.
  }
  \label{fig: vis_effect}
\end{figure}

\paragraph{Effectiveness of Each Instruction} Our video generation approach incorporates three distinct instructions: text, first frame, and last frame instruction. In this section, we will probe deeply into the influence of each instruction on the quality of generated videos.

\begin{figure}[t]
  \centering
  \includegraphics[width=0.475\textwidth]{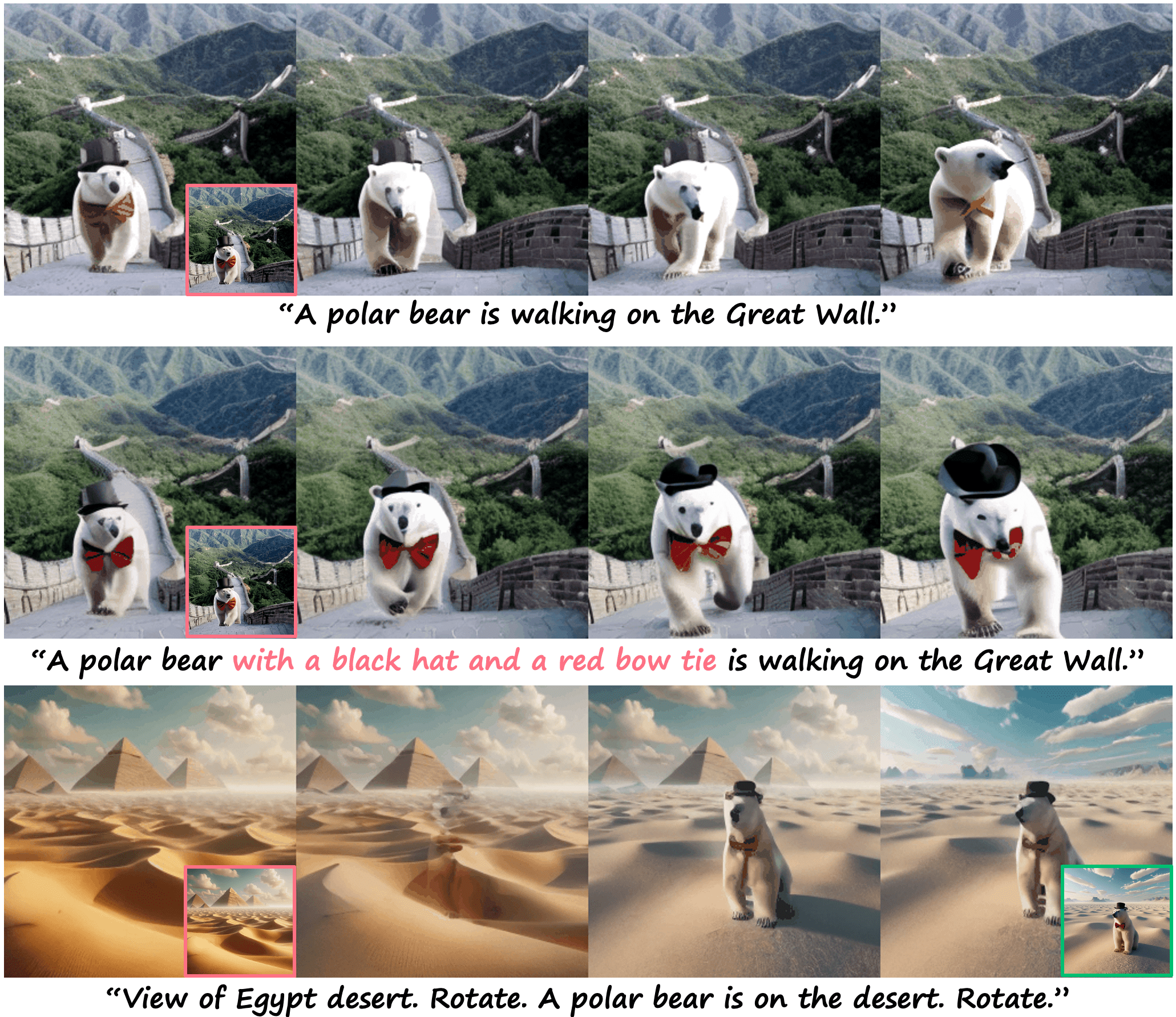}
  \caption{
  First two rows: text instruction helps enhance the cross-frame consistency of key elements like the \textit{black hat and red bow tie} of the polar bear. Last row: natural shot transition. 
  }
  \label{fig: bear_great_wall}
\end{figure}

In \ours, the text instruction could be concise, considering that the first frame instruction has delivered the objects/characters and scenes, which are challenging to be described succinctly and precisely with text. Nonetheless, the text prompt plays a vital role of specifying various motions, including but not limited to body movements, facial expressions, object movements, and visual effects (first two rows of Figure~\ref{fig: vis_rotate}). Besides, it allows for manipulating camera movements with specific prompts like "zoom in/out," "rotate," and "close-up," as demonstrated in the last row of Figure~\ref{fig: vis_rotate}. Moreover, the text instruction helps to hold the cross-frame consistency of specified key elements, such as the detailed descriptions of characters (polar bear in Figure~\ref{fig: bear_great_wall}).

The first frame instruction significantly improves the video quality by providing finer visual details. Moreover, it is key to generate multiple consecutive video clips. With the text and first frame instructions, \ours~is able to generate more motion-rich videos (Figure \ref{fig: vis_rotate} and Figure \ref{fig: bear_great_wall}) \tcb{compared to existing models.}

The last frame instruction, delineating the concluding status of a video clip, provides an additional control on video generation. This instruction is instrumental for synthesizing intricate motions, and becomes particularly crucial for out-of-domain video generation as depicted in the first two samples in Figure \ref{fig: teaser} and Figure \ref{fig: vis_effect}. 
Furthermore, we can generate a natural shot transition using last frame instruction (last sample of Figure~\ref{fig: bear_great_wall}).

\begin{figure}[t]
  \centering
  \includegraphics[width=0.465\textwidth]{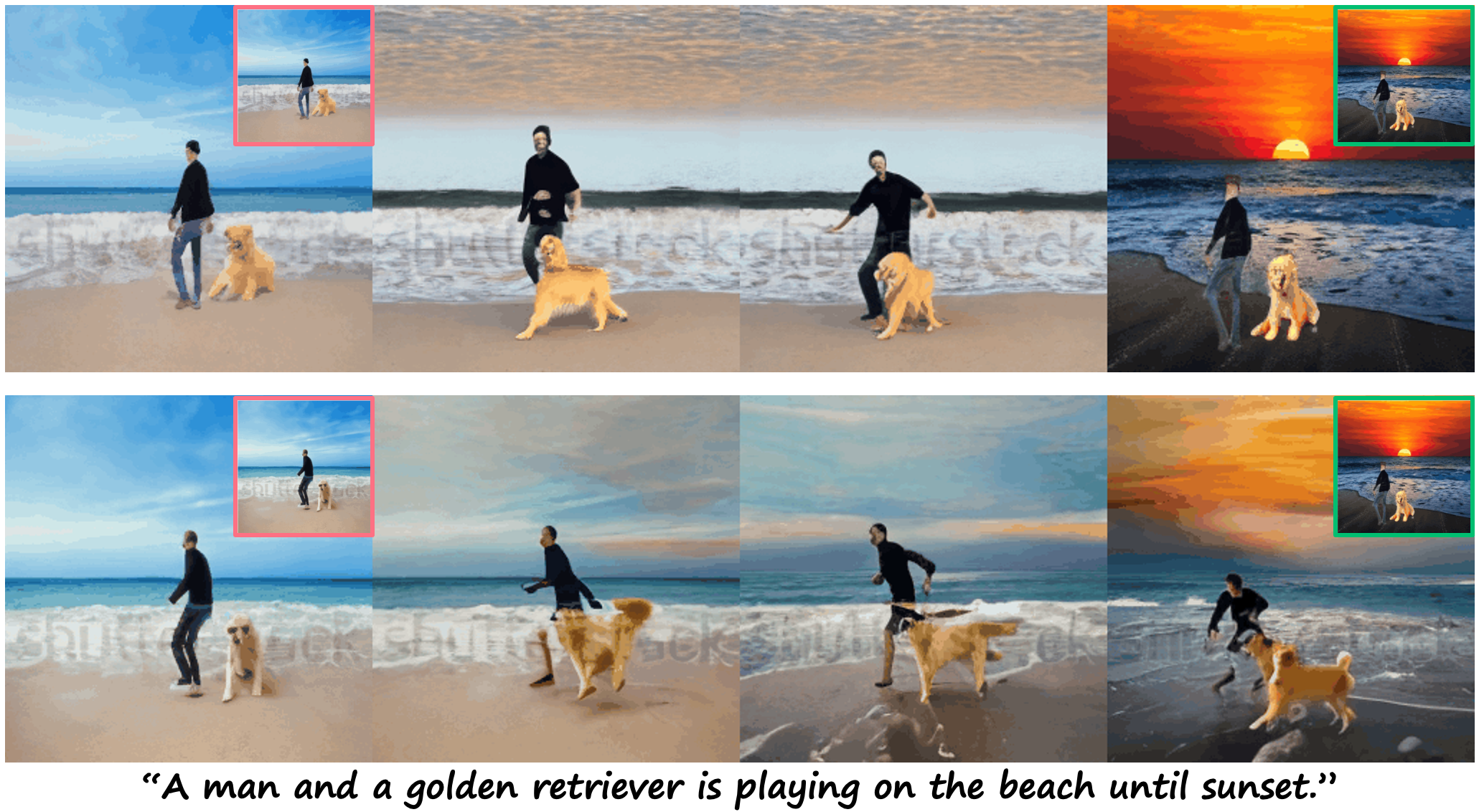}
  \caption{
  Illustration of the effectiveness of the proposed techniques ($\tau=25$) to avoid replicating the last frame instruction. 
  }
  \label{fig: man_dog_beach}
\end{figure}

\paragraph{Strength of Last Frame Guidance}
To make the model work well with user-provided drafts, even if they are somewhat imprecise, we intentionally avoid encouraging the model replicate the last frame instruction exactly, with the proposed techniques detailed in Sec. \ref{sec: method}. As shown in the Figure~\ref{fig: man_dog_beach}, without our techniques, the generated video abruptly ends in the given last frame instruction exactly. In contrast, with our proposed methods, the generated video is more fluent and temporally consistent.

\paragraph{Generalization to Out-of-Domain Image Instructions} 
\tcb{Despite the notable lack of training videos in non-realistic styles (\textit{e.g.}, science fictions, comics, and cartoons), \ours~demonstrates a remarkable capability to generate high-quality videos in these out-of-domain categories.} This generalizability can be attributed to that our approach focuses on learning dynamics and ensuring temporal consistency, given the image instructions. As \ours~learns the underlying principles of motions in real world, it can generalize across various stylistic domains of image instructions.

\subsection{Ablation Study}
\label{sec: ab}

\begin{table}[ht]
\begin{center}
\small
\captionof{table}{Ablation study results on UCF-101.}
\begingroup

\setlength{\tabcolsep}{20pt}
\begin{NiceTabular}{@{}lcc@{}}

\toprule
Method &
FID($\downarrow$) &
FVD($\downarrow$) \\

\midrule
\ding{192} \tv~baseline & 59.35 & 450.58 \\
\midrule
\ding{193} \ours & 49.36 & 242.82 \\
\ding{194} \ours~w/o $\mathbf{c }^{text}$ & 51.26 & 375.79 \\
\ding{195} \ours~w/o $\mathbf{f}^{last}$ & 49.45 & 339.08 \\
\bottomrule

\end{NiceTabular}

\endgroup
\label{table: ablation}
\end{center}
\end{table}

\begin{figure}[t]
  \centering
  \includegraphics[width=0.48\textwidth]{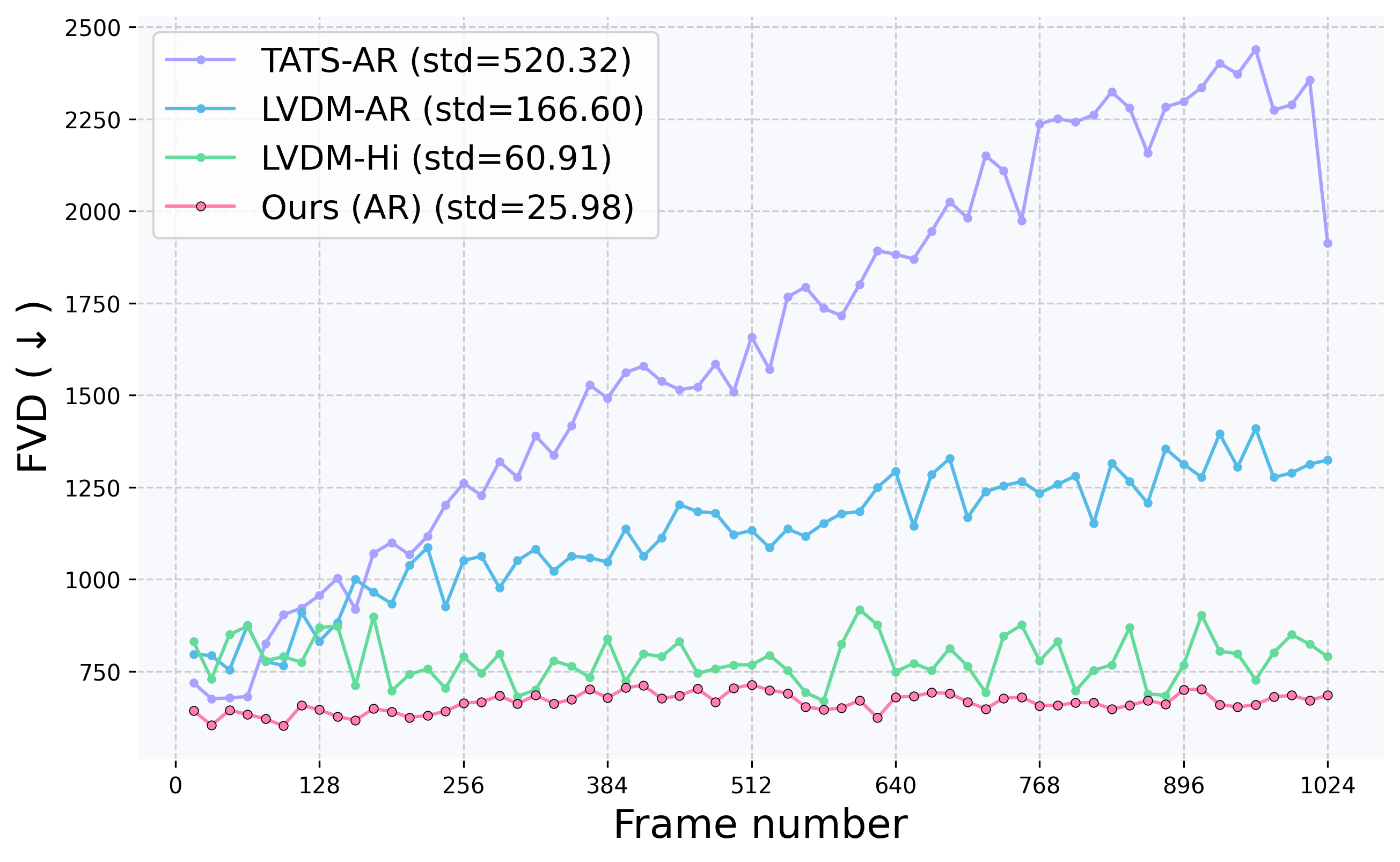}
  \caption{
  FVD comparison for long video generation (1024 frames) on UCF-101. AR: auto-regressive. Hi: hierarchical. \tcb{The construction of long video with \ours~is in an autoregressive manner.}
  }
  \label{fig: ucf_long}
\end{figure}

To evaluate the key components of \ours, we conduct a quantitative ablation study on the UCF-101 dataset following the zero-shot evaluation setting in Sec. \ref{sec: quant_eva}.

First, we provide a \tv~baseline (\ding{192}) for comparison trained on the same dataset. We further analyze the effectiveness of instructions employed in our model. Given the indispensable nature of the first frame instruction for the generation of continuous video clips, our ablation focuses on the text instruction (\ding{194}) and the last frame instruction (\ding{195}). The experimental results indicate that omitting either instruction results in a significant deterioration in video quality. Notably, even though the evaluation does not incorporate the last frame instruction, model trained with this instruction (\ding{193}) outperforms the model trained without it (\ding{195}).
This observation \tcb{reveals} that relying solely on the \texttt{<text, first frame>} for video generation poses substantial challenges due to the \tcb{significant diversity of video content.} In contrast, incorporating all three instructions enhances \ours’s capacity to model motion dynamics and hold temporal consistency.

\begin{figure*}[t]
  \centering
  \includegraphics[width=0.99\textwidth]{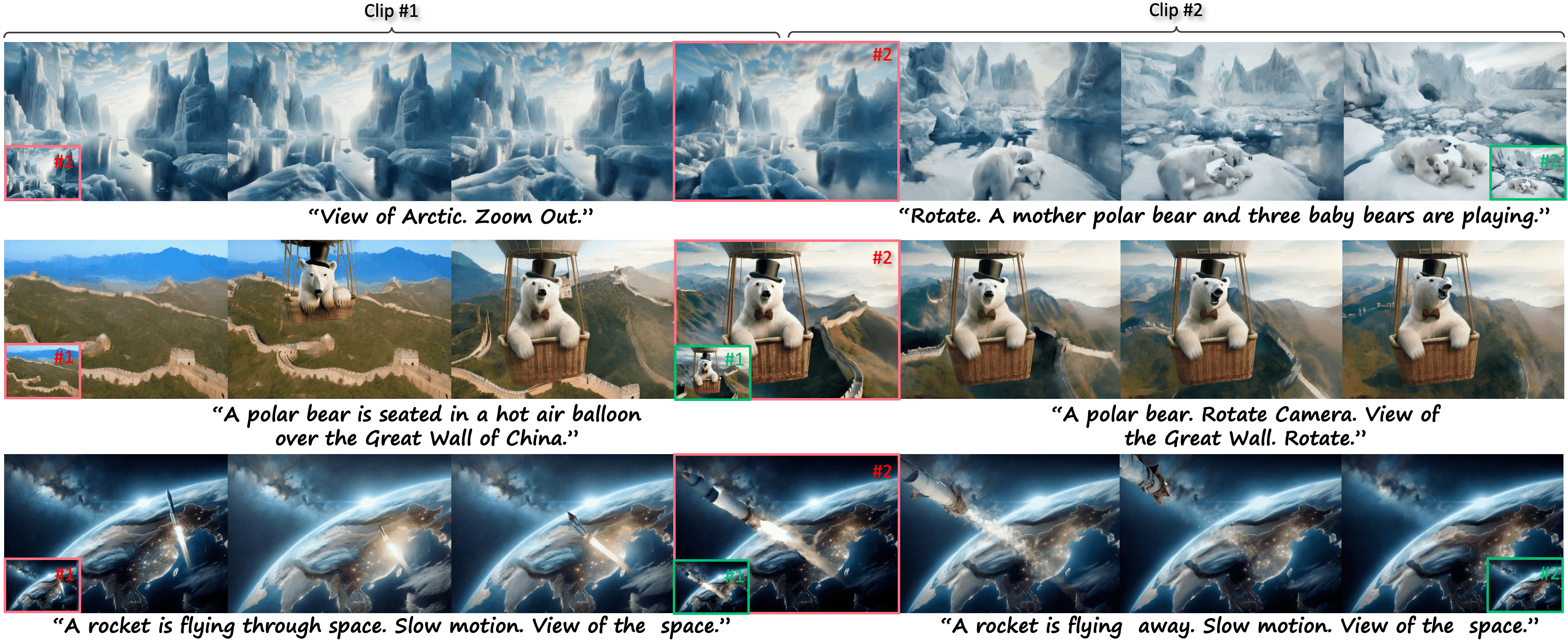}
  \caption{
  Illustration of \ours~handling intricate shot compositions consisting of two continuous video clips, in which case the last frame of the Clip \#1 serves as the first frame instruction for Clip \#2.
  }
  \label{fig: bear_long}
\end{figure*}

\subsection{Long Video Generation}
\label{sec: exp_long}

\subsubsection{Quantitative Evaluation}
As aforementioned, \ours~is trained to strictly adhere to the first frame instruction, in order to generate long videos, where the last frame from preceding clip is used as the first frame instruction for generating the subsequent clip. To evaluate \ours's capability of long video generation, we follow the previous work \cite{ge2022TATS,he2022LVDM} and generate 512 videos with 1024 frames on UCF-101 datasets, \tcb{under the zero-shot setting detailed in Sec. \ref{sec: quant_eva}}. We report the FVD of every 16 frames extracted side-by-side from the synthesized videos. The results, as shown in Figure~\ref{fig: ucf_long}, show that \ours~demonstrates lower FVD scores and smoother temporal variations, compared with auto-regressive models, TATS-AR \cite{ge2022TATS} and LVDM-AR \cite{he2022LVDM}, and the hierarchical approach LVDM-Hi. Please refer to the Supplementary for visual comparisons.

\begin{figure}[t]
  \centering
  \includegraphics[width=0.475\textwidth]{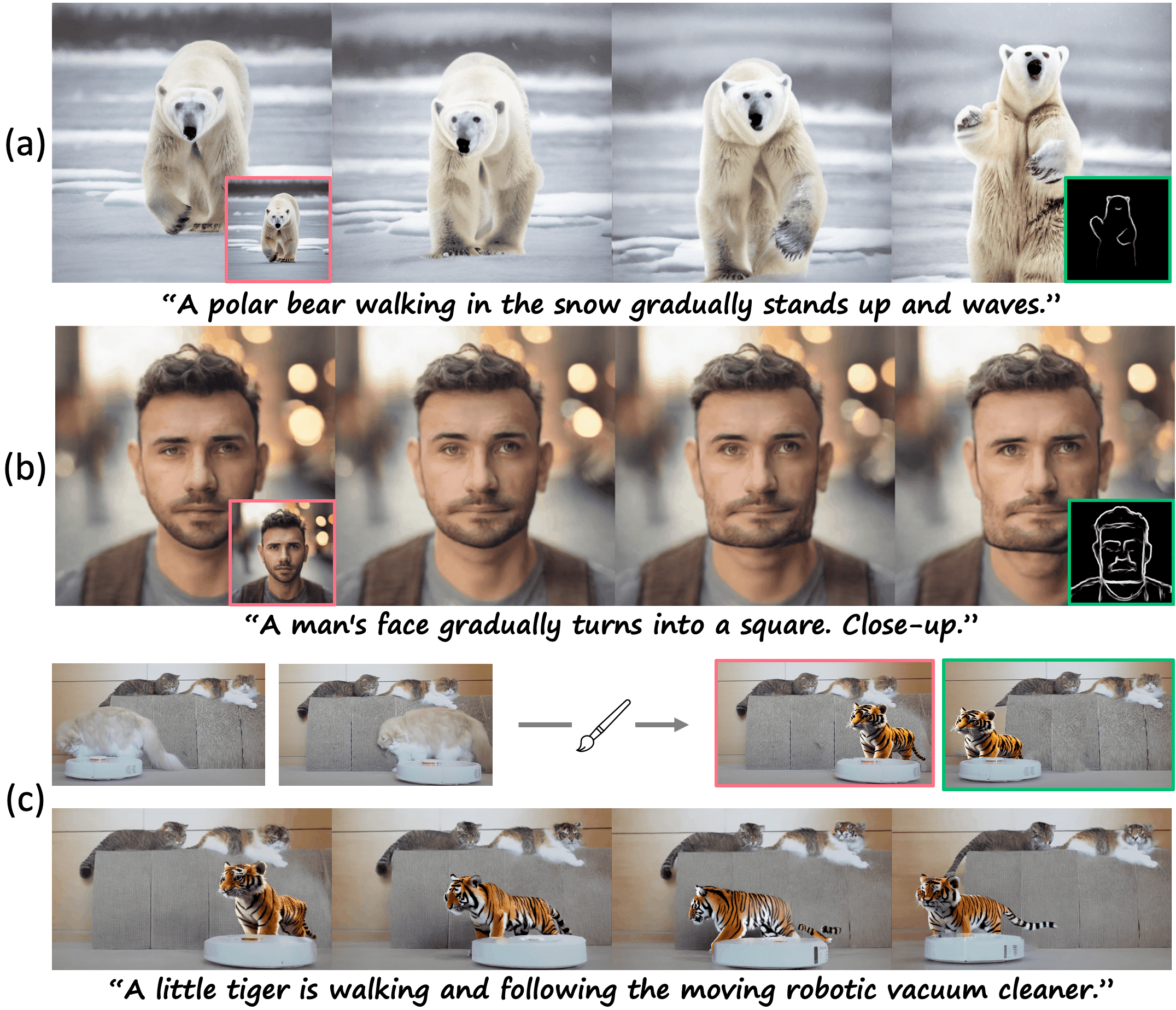}
  \caption{
  Illustration of video generation with sketch image as last frame instruction (first two examples), and \ours~for zero-shot video editing (c). 
  }
  \label{fig: more_app}
\end{figure}

\subsubsection{Qualitative Analysis} 
Recognizing that most real-world long videos (\textit{e.g.}, videos or films on YouTube) comprise numerous shots rather than a single continuous shot, this qualitative analysis focuses on \ours's capabilities of generating a composite shot. This is formed by stringing together multiple continuous video clips that are temporally consistent. Figure~\ref{fig: bear_long} illustrates the capability of \ours~to handle intricate shot compositions involving complex camera movements (in Arctic scenes), smooth animation effects (polar bear appears in a hot air balloon over the Great Wall), and precise control over the trajectory of a rocket. These instances exemplify how users interact with \ours~to craft desired video sequences. \tcr{Leveraging \ours's advanced generation capabilities, we have successfully synthesized a three-minute video that not only tells a coherent story but also maintains a consistent portrayal of the main character. }

\subsection{More Applications}
\label{sec: more_application}

\paragraph{Sketch Instruction}
Our proposed approach can be extended to other types of image instructions, such as semantic maps, image sketches, human poses, and bounding boxes. To demonstrate this, we take the image sketch as an example and finetune \ours~with image sketch \cite{xie2015HED} as the last frame instruction. The results are shown in the first two rows of Figure \ref{fig: more_app}, exhibiting that a simple sketch image is able to guide the video generation process.

\paragraph{Zero-shot Video Editing}
\ours~is able to perform video editing without any training, achieved by transforming the video editing task into an image editing task. As shown in the last example in Figure \ref{fig: more_app}, by editing the first frame and the last frame of the provided video, \ours~generates a temporally consistent video aligned with user expectation on video editing.

\section{Conclusion}

In this paper, we proposed a novel video generation approach based on diffusion models, \ours, which incorporates image instructions for both the first and last frames in conjunction with text instruction. We developed training and inference techniques tailored for this approach. \ours~trained \tcb{mainly} on WebVid-10M exhibited exceptional proficiency in synthesizing videos with complex scenes and actions, setting a new standard in video generation.

While our approach has achieved noteworthy results, there is potential for further advancements. First, the model can benefit from training with high-quality, open-domain video data. Second, 
fine-tuning the model within specific domains could further augment its capabilities. Third, incorporating
annotated texts that outline key elements and motions of videos could improve the alignment to user's instructions.
Lastly, \ours~currently consists of only 1.5B parameters, presenting an opportunity for future scaling up. 
Further investigation into these aspects will be explored in future work.

{
    \small
    \bibliographystyle{ieeenat_fullname}
    \bibliography{ref}
}

\end{document}